\documentclass[10pt,twocolumn,letterpaper]{article}

\usepackage{wacv}
\usepackage{times}
\usepackage{epsfig}
\usepackage{graphicx}
\usepackage{makecell}
\usepackage{amsmath}
\usepackage{amssymb}
\usepackage{booktabs}

\wacvfinalcopy 

\def\wacvPaperID{571} 

\ifwacvfinal\fi
\begin{document}

\title{Autonomous Driving in Reality with Reinforcement Learning and Image Translation}

\author{Nayun Xu \hspace{2cm} Bowen Tan \hspace{2cm} Bingyu Kong \\
Shanghai Jiao Tong University\\
{\tt\small \{xunayun, tanbowen, crystal\_mis\}@sjtu.edu.cn}
}

\maketitle
\ifwacvfinal\thispagestyle{empty}\fi

\begin{abstract}
Supervised learning is widely used in training autonomous driving vehicle. However, it is trained with large amount of supervised labeled data. 
Reinforcement learning can be trained without abundant labeled data, but we cannot train it in reality because it would involve many unpredictable accidents. Nevertheless, training an agent with good performance in virtual environment is relatively much easier. Because of the huge difference between virtual and real, how to fill the gap between virtual and real is challenging.
In this paper, we proposed a novel framework of reinforcement learning with image semantic segmentation network to make the whole model adaptable to reality. The agent is trained in TORCS, a car racing simulator.
\end{abstract}

\begin{figure*}[!t]
\centering
\includegraphics[width=\textwidth]{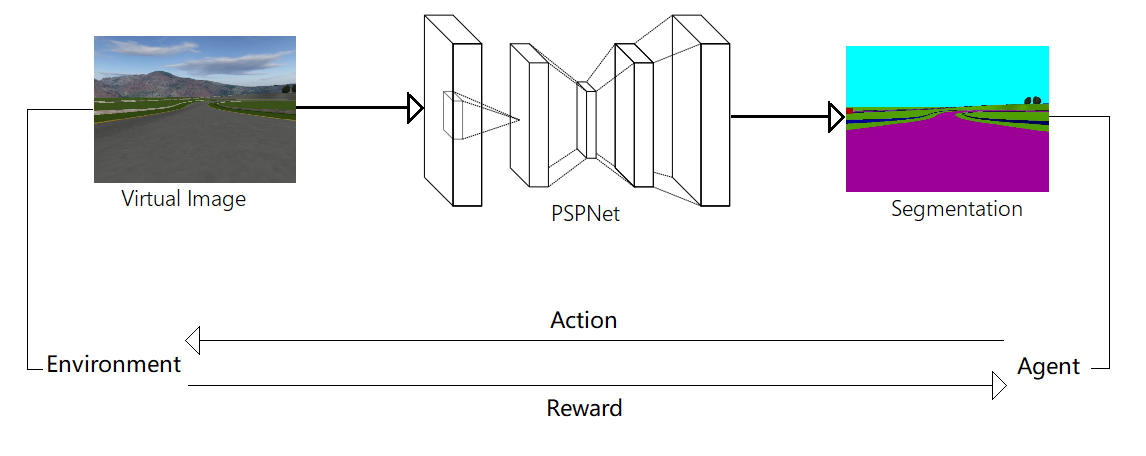}
\caption{The reinforcement learning agent is trained as following: Initially, a virtual image is produced by simulator TORCS.Then this image go through an image semantic segmentation network. Finally the semantic representation of current driving scene is fed into the reinforcement learning agent.
In reinforcement learning framework, agent observes the semantic image and chooses an action. After agent taking action, the environment will give a reward to agent to help the agent adjust its parameters and give an image of next state to the semantic segmentation network}
\label{fig:pipline2}
\end{figure*}

\section{Introduction}

In the artificial intelligence field, autonomous driving is a significant task and closely relevant to computer vision. The intention of this task is to design a system to autonomously control vehicles to do actions, such as steering, accelerating, and braking. Essentially, autonomous driving is an interactive model with the environment. In this task, on one hand, computer vision techniques can help the system extract and analyze information from driving scene, On the other hand, this task also requires many other techniques to do the decision making.
   
There are mainly two categories of methodologies to deal with this task, supervised learning and reinforcement learning. Both methods face obstacles. When using supervised learning to do the autonomous driving, the most crucial problem is training relies on a large amount of labeled data which requires much human effort. Also, it is hard to develop an end-to-end model because the ground truth for action is not objectively determined and the label on action would include personal bias. And state-of-the-art methods not using an end-to-end style are mostly very complex involving many empirical rules. Therefore, reinforcement learning seems better fits this task because it can be trained without human labeled training data. Reinforcement learning avoids the problem of a large amount of labeled data and the potential shortcoming of human bias. 
	
Nevertheless, reinforcement learning also meets many problems. The most fundamental problem is that reinforcement learning model cannot be trained in reality because the training process would involve many collisions and other unpredictable situations. So most reinforcement learning models on this task are trained in virtual simulators. The approach of training also brings about some problems. The model’s performance when it is applied in reality will largely depend on how real the simulator is. In other words, the real driving scene is more complicated than the virtual simulator, which also challenges the generalization ability of the model. 
    
In this paper, our approach is focused on how to fill the gap between virtual and real. We want to solve mainly two problems. One is the difference between the virtual simulator and real in terms of driving scenes. The other is the complexity and noise of reality scenes. Therefore, we use a translation network to transfer the virtual driving scene to semantic segmentation image and use these semantic images as state input to the agent. When applying the model to reality, we do semantic segmentation on the real driving scene and use the segmentation result as input to our model. We consider this translation would fill the gap between virtual and real. Also, we consider the semantic segmentation would be an appropriate level of abstraction of the real driving scene which reduces the complexity and still holds most useful information such as lanes and barriers. 

Our framework has several advantages as below:
\begin{itemize}
\item Compared with the huge demand for labeled data in state-of-art supervised learning, our framework does not relies on any labeled data.
\item Training in a virtual environment and transfer to the real world, we do not need to confront the danger and enormous loss of failure.
\item The input of the reinforcement learning agent is semantic segmentation image. Semantic image contains less information compared to original image, but includes most information needed by agent to take actions. In other words, semantic image neglects useless information in original image.    
\end{itemize}
\begin{figure*}[!t]
\centering
\includegraphics[width=\textwidth]{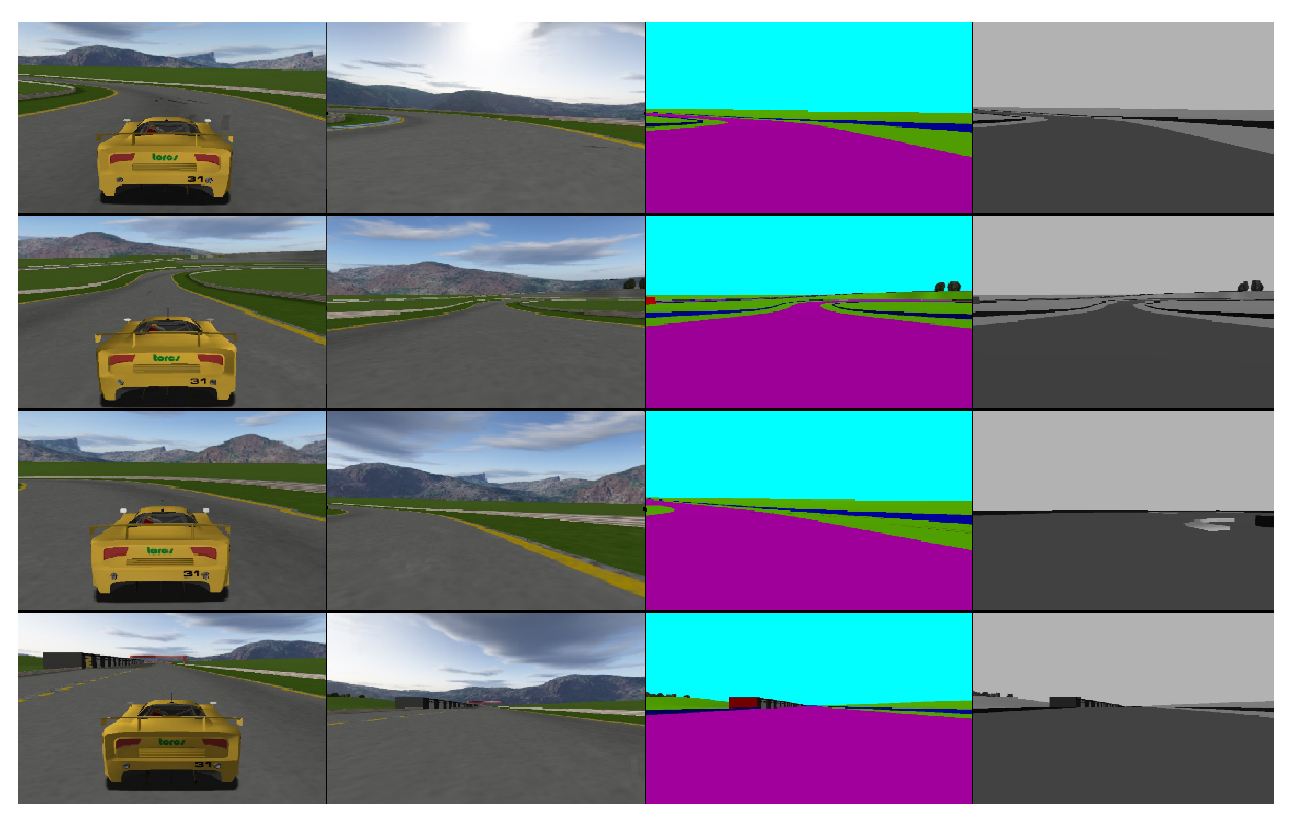}
\caption{Examples of translation from the original output got from simulator to our intended input for the reinforcement learning agent. The first column is the original scene displayed by TORCS, the second column is the corresponding first-person perspective scene got by hacking from the source code, the third column is the semantic segmentation of the first-person perspective scene, the fourth column is the gray scale semantic perspective.}
\label{fig:virtual-parse}
\end{figure*}

\section{Related Work}
\subsection{Supervised Learning for Autonomous Driving}
Supervised learning has been used in autonomous driving for decades. And these work can be categorized into two major styles, perception-based approaches and end-to-end approaches. The perception-based approaches detect some mediate information to help the agent make decisions. In early years, they detect driving-relevant objects such as lanes, cars, pedestrians, etc. Recently, approaches like deep driving involve CNNs~\cite{CNNs} to detect more direct information like distances between cars and lanes. 

The end-to-end approaches seem more direct. They want to directly map images input to driving action predictions. ALVINN \cite{pomerleau1989alvinn} shows an early attempt of end-to-end approaches. It learns the direct mapping with a shallow neural network. And in recent years, the shallow network has been replaced by more powerful deep neural networks like CNNs. NVIDIA \cite{nvidia}recently presented an end-to-end system with deep learning methods~\cite{EndtoEndLearningforSelf-DrivingCar}. 

However, whichever styles of supervised learning approaches are employed, the training process requires large quantities of labeled data. And the model performance is highly relevant to the quality and quantity of data. Besides, supervised models have limited generalization abilities because the real driving environments are numerous which are far beyond the training data.

\subsection{Reinforcement learning for Autonomous Driving}
With a lot of variations, reinforcement learning has been a common technique for many scenarios such as computer games \cite{mnih2015human} and robot control\cite{kohl2004policy,endo2008learning}. Recently, plenty of work \cite{abbeel2007application, 
DBLP:journals/corr/Shalev-ShwartzS16a} contributed to building an autonomous driving system with reliable security.
However, high-dimensionality of state space and non-trivial large action range in the real world's practical driving environments are challenging the training of reinforcement learning. 
It is time-consuming to get an optimal policy over such high complexity.
With the power of deep neural networks and deep reinforcement learning\cite{koutnik2013evolving, mnih2015human, schulman2015trust,lillicrap2015continuous,A3C}, a great step forward in such complexity is made recently.
Nonetheless, not only deep Q-learning\cite{mnih2015human} method but policy gradient method\cite{lillicrap2015continuous} , they both require the interaction between the agent and environment to get feedback and reward.
Obviously, training agents of an autonomous vehicle in real-world scenes is unrealistic because of the huge cost for every wrong action.

In order to avoid damage to the real world, reinforcement learning with driving simulators and transfer learning models appear. The training process of reinforcement learning with a driving simulator is safe and fast. However, in order to drive autonomously in the real world, the driving agent must be able to take actions according to an intricate visual field, which is much different from the virtual images we get from a driving simulator. Models trained on virtual data and simulator cannot perform well in real-world data. 

For the past decade, there are many models \cite{DBLP:journals/corr/RusuVRHPH16,
gupta2017learning,tobin2017domain}. These models either first 
train a model in virtual environment and then fine-tune in the 
real environment \cite{DBLP:journals/corr/RusuVRHPH16}, or learn 
an alignment between virtual images and real images by finding 
representations that are shared between the two domains 
\cite{tzeng2016adapting} contribute to the transferring reinforcement learning. Some of these models first trained on virtual data to reduce time training in real-world significantly. But these models still have to train in real-world, they cannot avoid the risk of damage to real-world radically. Some other models try to learn an alignment between the virtual image and real image. In reality, the real visual field is much more complicated and more noisy than virtual image. Challenge of these models are, under a certain condition, an alignment which is good enough cannot be found between virtual images and real images. 

There is a recent work \cite{DBLP:journals/corr/YouPWL17} managing to train the reinforcement learning only in environment created by simulator TORCS\cite{wymann2000torcs}. This novel framework demonstrated that after training on a simulator, the autonomous agent is able to realize a collision-free flight. Nonetheless, the work needs a nontrivial training environment to achieve its goal.

\subsection{Scene Parsing}
Semantic image segmentation is one part of our model. It can be seen as a pixel-level prediction task. Based on a deep convolutional neural network and fully convolutional neural network\cite{Long_2015_CVPR}, many works achieved good performance in the field of image segmentation\cite{badrinarayanan2015segnet}. What we used in our framework to do image segmentation is PSPNet\cite{DBLP:journals/corr/ZhaoSQWJ16}. It extends the pixel-pixel level feature to specially designed global pooling one. And it also proposes an optimization strategy with a deeply supervised loss. This work achieves state-of-the-art performance on various datasets.

\section{Proposed Framework}
Our goal is to develop an autonomous driving model trained entirely in a virtual environment which can be applied in real-world driving scenes with good performance. One of the major challenges is that the training environment is generated by a simulator, which means the training environment would be quite different from real-world scenes in terms of their appearance. To tackle this problem, we proposed to use an image translation process to convert virtual images to semantic layouts which resembles the semantic segmentation of its supposed corresponding real world image. This idea is inspired by the work of \cite{DBLP:journals/corr/YouPWL17} which tries to fill the gap between virtual and real on synthesized images. Our framework contains two parts, the image translation process and the  reinforcement learning. The image translation process intends to translate the virtual driving scene to a semantic representation. We use PSPNet as the semantic segmentation network in our model. In this part, in order to get required information such as the first-person perspective from the TORCS simulator, we use some hacking techniques. The sample process of this translation part is presented in Figure \ref{fig:pipline2}. Finally, we train an autonomous driving car using reinforcement learning on the semantic layouts obtained by the translation network. In the reinforcement learning part, we use the asynchronous advantage actor-critic reinforcement learning algorithm. In this section, we will present the image translation process and how to apply reinforcement learning to train an autonomous driving agent.

\subsection{Image Translation Process}
 
 In order to ensure our autonomous model entirely trained in a virtual environment has good performance in the real world, we have to fill the gap between the training environment generated by simulators and the real-world scenes. From our point of view, the semantic segmentation of real-world visual field contains enough information the agent needs to take driving actions. Inspired by the work of \cite{DBLP:journals/corr/YouPWL17}, we adopt a translation process to translate the virtual image into a semantic segmentation image. This translation is based on the hypothesis that the semantic segmentation of real-world visual field is similar to the segmentation of a virtual image.

The image translation process mainly includes an image segmentation network. 

\begin{figure}[!h]
\includegraphics[width=0.5\textwidth]{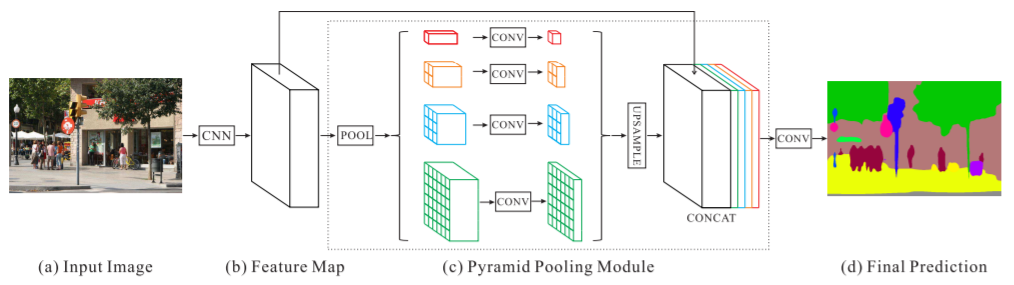}
\caption{The convolutional neural network in PSPNet is used to get the feature map from the last convolutional layer. After that, in order to harvest different sub-region representation, the PSPNet applies a pyramid parsing module. Then, it uses upsampling and concatenation layers to get the final feature representation. Finally, the convolution layer, which gets the final feature representation as input, outputs per-pixel prediction.}
\label{fig:PSPNet}
\end{figure}

First, we get the first-person perspective driving scene from the TORCS simulator. Then we use PSPNet to translate the virtual driving image into semantic segmentation image. The output of this part is the semantic layout, and the semantic layout will be fed into the reinforcement learning agent.

There is an obvious obstacle in this part. The appearance of virtual driving images are different from the real-world images, we cannot apply the segmentation tool pretrained on real-world datasets like Cityscapes\cite{DBLP:journals/corr/CordtsORREBFRS16}  directly to the virtual images. And there are no semantic annotations for TORCS virtual images. We tackle this problem by some hacking techniques which will be explained in the experiment part.

\subsection{Reinforcement Learning for Training Autonomous Driving}

\begin{figure}[!h]
\includegraphics[width=0.48\textwidth]{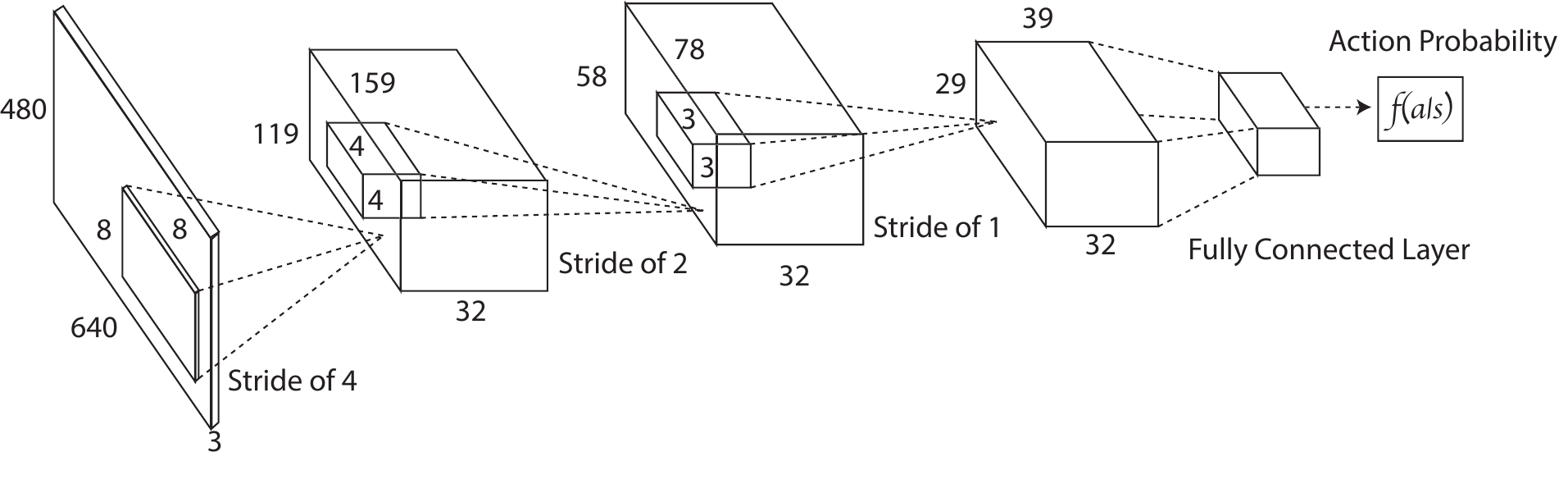}
\caption{Reinforcement learning network architecture: an end-to-end network mapping state representations to action probability outputs.}
\label{fig:rl_network}
\end{figure}

We use Asynchronous Advantage Actor-Critic(A3c) to train the autonomous-driving vehicle and get the best performance comparing with other reinforcement learning structures,
which is a conceptually simple and
lightweight framework for deep reinforcement
learning that uses asynchronous gradient
descent for optimization of deep neural network
controllers.
A3C utilizes multiple incarnations of the above in order to learn more efficiently.

In A3C there is a global network and multiple worker agents which each have their own set of network parameters. Each of these agents interacts with its own copy of the environment at the same time as the other agents are interacting with their environments.

The experience of each agent is independent of the experience of the others, making speed up and better performance.
In this way, the overall experience available for training becomes more diverse.
Critically, the agent uses the value estimate to update the policy more intelligently than traditional policy gradient methods.

\begin{figure}[!h]
\includegraphics[width=0.5\textwidth]{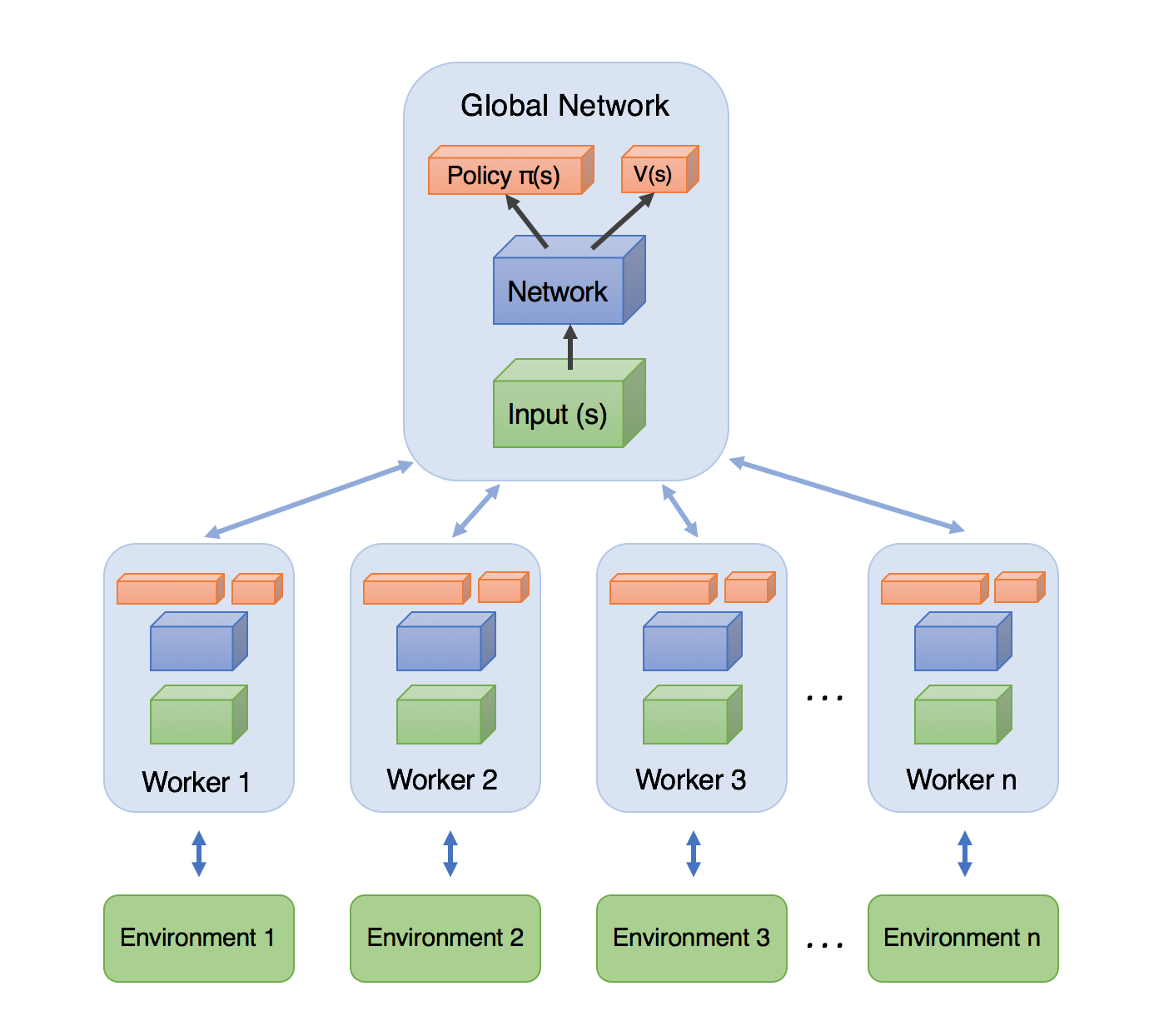}
\caption{Diagram of A3C high-level architecture.}
\label{fig:a3c}
\end{figure}

In \cite{A3C} there are more details of implement A3C algorithm.

In order to encourage the agent to drive faster and avoid collisions, 
we define the reward function as
\begin{align}
	r_t = \left\{
	\begin{array}{ll}
		(v_t \cdot \cos\alpha - \mathrm{dist}_{\text{center}}^{(t)})\cdot \beta & \text{no collision,}\\
		\gamma & \text{collision},
			\end{array}
			\right.
			\label{eq::reward}
		\end{align}
where $v_t$ is the speed (in $m/s$) of the agent at 
time step $t$, $\alpha$ is the angle (in rad) between 
the agent's speed and the tangent line of the track, 
and $\mathrm{dist}_{\text{center}}^{(t)}$ is the distance 
between the center of the agent and the middle of 
the track. $\beta, \gamma$ are constants and are determined
at the beginning of training. We take $\beta = 0.006, 
\gamma = -0.025$ in our training.

\section{Experiments}

We perform experiments to compare the performance of our method and other existing methods of autonomous driving on real-world driving data. This set of experiments aim to evaluate our model's performance in the real-world driving scene. And also we perform experiments to compare our method and basic reinforcement learning without image translation network on TORCS simulator. This set of experiments aim to show advantages our model bring to reinforcement learning process.

\subsection{Autonomous Driving with Image Translation Network and RL on Real-world Driving Data}\label{exp1}
In this experiment, we trained our proposed reinforcement 
learning model with image translation process. We first trained the semantic segmentation network(PSPNet) and then apply the trained network to generate semantic parsing images to feed into our A3C reinforcement learning agent to train a driving policy. And finally, we apply the trained agent on a real-world driving data to evaluate its performance. When we apply the agent to the real-world driving data, we also use the semantic segmentation network to get the semantic input to the agent.
To have a comparison, we also trained another reinforcement learning method without image translation network in the TORCS simulator. This model is same as our proposed model except the translation network. We call it Basic RL. 

\subsubsection{Dataset}
The real-world driving data is from \cite{dataset}, which is collected with detailed steering angle autonomous per frame. There are in total around 45k images in the dataset. To train the image translation network, we use two datasets separately. We collect 2k images from the TORCS simulator and use hacking techniques to get semantic labels for these collected virtual images. We modified the source code of TORCS guided by the work to control each category of objects to show or hide. In detail, we compare the original image with the image after we hide all the trees, and get the exact pixels covered by trees. And we do the same for other objects. We process the collected images in this way to get their semantic labels. The other dataset we use is Cityscape\cite{DBLP:journals/corr/CordtsORREBFRS16}. It contains around 25k real-world images with semantic segmentation annotations.

\begin{figure*}[!t]
\includegraphics[width=1\textwidth]{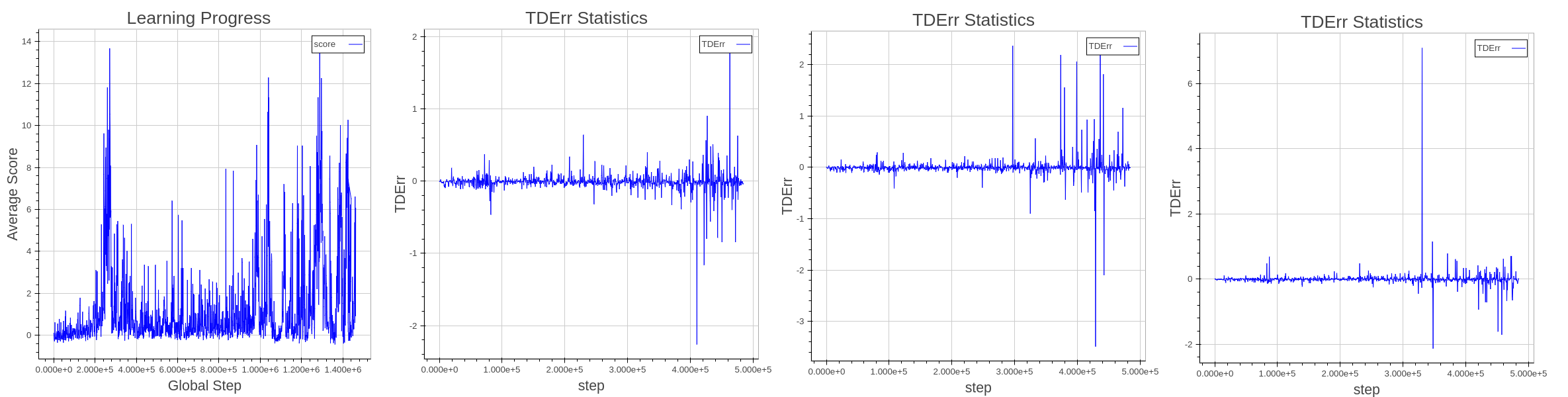}
\caption{The training process statistics of RGB observation.}
\label{fig:rgb}

\end{figure*}
\begin{figure*}[!ht]
\includegraphics[width=1\textwidth]{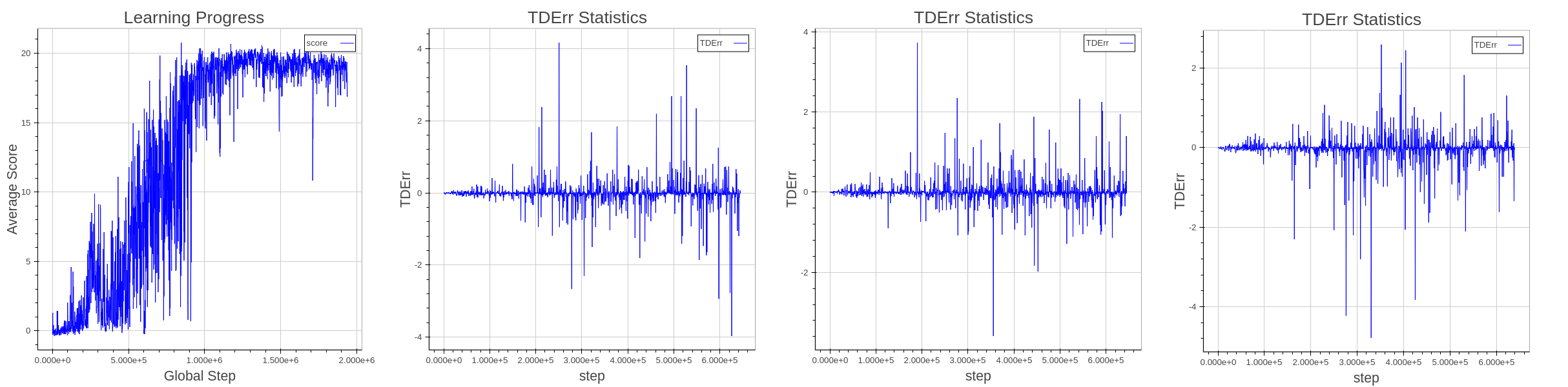}
\caption{The training process statistics of gray scale observation.}
\label{fig:gray}
\end{figure*}

\subsubsection{Image Translation Process}
Both scene segmentation parts to translating virtual images and real-world images in our framework are PSPNet.

We train the network translating virtual images to semantic ones which would be applied in training part with the dataset we collected from TORCS. We don't simply use the ground truth label we hacked from the simulator as the output of this part because we want to avoid the bias between PSPNet result and hacked ground truth. More specifically, we want the training part in TORCS and testing part on real-world data share more similarity through using same segmentation tool for both parts.

And we train the network translating real-world images to semantic ones which would be applied in real-world testing part with the Cityscape dataset.

We do experiments to determine whether use the original RGB semantic result as the output of this part and feed it into the agent or translate the semantic result into a gray scale image before fed into the agent. We finally choose to use the gray scale images as the output of this part. This will be presented in the result section.

\subsubsection{Reinforcement Training}
We use such structure in our A3C algorithm: the actor network is a 4-layer convolutional network, with ReLU as activation function. It's input is 4 consecutive frames and there are 9 discrete actions can be its output(``go straight with acceleration'', ``go left with acceleration'',
``go right with acceleration'', ``go straight and brake'', ``go left
and brake'', ``go right and brake'', ``go straight'', ``go left'', and
``go right''). 12 asynchronous threads with RMSProp optimizer is what we used to train our reinforcement agent, whose initial learning rate is $0.01$, $\gamma=0.9$ and $\epsilon=0.1$.

\subsubsection{Evaluation on Real-World Dataset}
The real world driving dataset \cite{dataset} provides the steering 
angle annotations per frame. 
However, the actions performed in the TORCS virtual environment only contain "going left", "going right", and "going straight" and their combination with "brake" and "acceleration". Therefore we come up with a mapping in Table \ref{tab:mapping} from the steering angle to the action space of our RL agent. With this mapping, we evaluate the accuracy of action prediction of our model.

\begin{table}[!htbp]
\centering
\caption{the steering angles and corresponding actions}\label{tab:mapping}
\begin{tabular}{ccc}
\toprule
angle(degree)& action\\
\midrule
$[-15, 15]$& going straight\\
less than $-15$& going left\\
more than $15$& going right\\
\bottomrule
\end{tabular}
\end{table}

\section{Result}
\begin{figure}[!hbtp]
\includegraphics[width=0.48\textwidth]{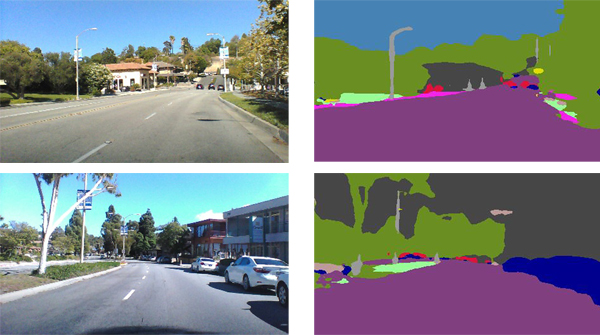}
\caption{Real images from the driving data and their semantic segmentations}
\label{fig:real-parse}
\end{figure}

\subsection{Result of Image Translation Process}
We use PSPNet to translate virtual images and real-world images into semantic images. Figure \ref{fig:virtual-parse}  is examples of the result of translation process on virtual images. We tried two alternative output of this part - the RGB image or gray scale image. We do experiments for both, compare the performance and finally choose the gray scale image.

Figure \ref{fig:real-parse} is examples of the translation result of real-world driving images.

\begin{table}[!htbp]
\centering
\caption{Accuracy of different models}\label{tab:comp}
\begin{tabular}{cccc}
\toprule
   & Our Model & Supervised Model & Basic RL\\
\midrule
Accuracy & $36.6\%$ & 52.6\% & 28.1\% \\
\bottomrule
\end{tabular}
\end{table}

\subsection{Result of Reinforcement Learning}
Figure \ref{fig:rgb}. shows the training process using RGB semantic image as agent observation.
Figure \ref{fig:gray}. shows the training process using gray scale semantic image as agent observation.

\subsection{Testing on Real-World Driving Data}
We extract 4 consecutive frames from the real-world driving data and parse them into semantic images with PSPNet. And feed these images into our agent to get predictions. Figure 5 shows the prediction our agent made to corresponding input.

We test all the frames in the driving data set and finally got an accuracy of 36.6\%. The comparison with the basic reinforcement learning and the supervised model trained on the dataset is shown in Table \ref{tab:comp}.

And examples of predictions and ground truth are shown in Table \ref{tab:mmp}.

\begin{table}[!htbp]
\centering
\newcommand{\tabincell}[2]{\begin{tabular}{@{}#1@{}}#2\end{tabular}}
\caption{The predictions on real-world driving data and their ground truth}\label{tab:mmp}
\begin{tabular}{ccccc}
\toprule
origin & semantic & prediction & label\\
\midrule
\includegraphics[scale=0.12]{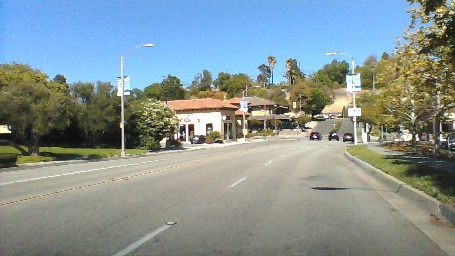}& \includegraphics[scale=0.12]{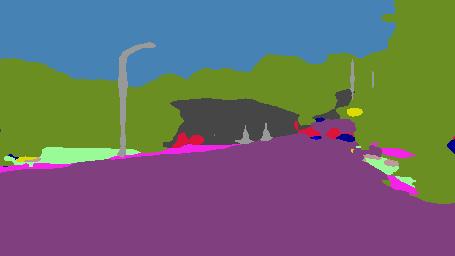} & \tabincell{c}{straight }  & $45.180000$\\
\midrule
\includegraphics[scale=0.12]{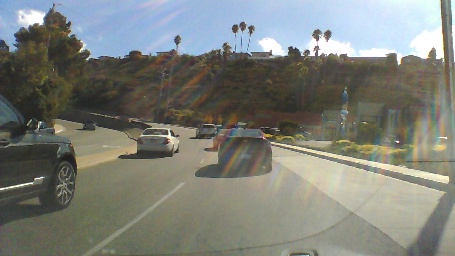}& \includegraphics[scale=0.12]{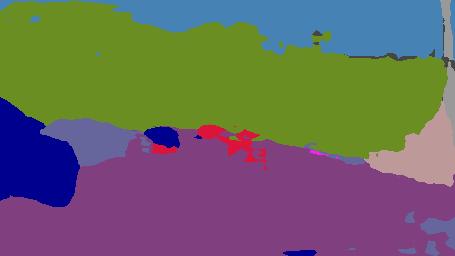} & \tabincell{c}{left} & $-12.400000$\\
\midrule
\includegraphics[scale=0.12]{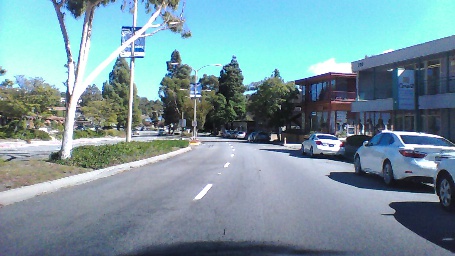}& \includegraphics[scale=0.12]{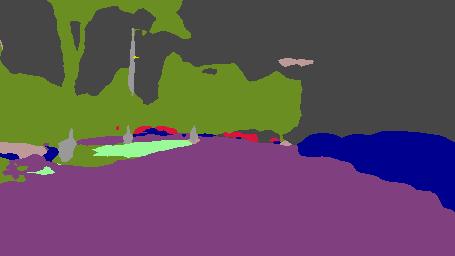} & \tabincell{c}{straight} & $-8.970000$\\
\midrule
\includegraphics[scale=0.12]{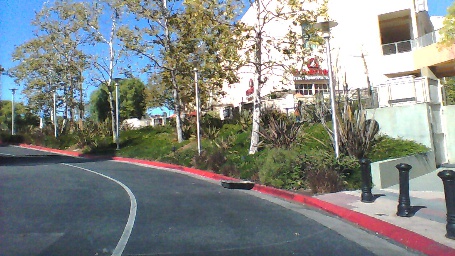}& \includegraphics[scale=0.12]{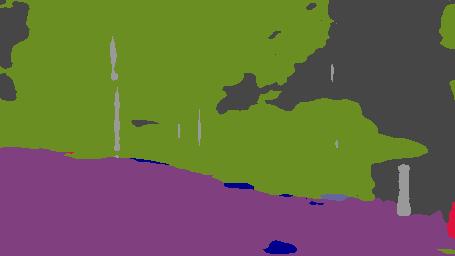} & \tabincell{c}{left} & $-95.700000$\\
\bottomrule
\end{tabular}
\end{table}

\section{Conclusion and Future Work}
Our proposed autonomous driving model tries to transfer the reinforcement learning agent developed in a virtual environment to real-world tasks. We use semantic segmentation as the tool to fill the gap between virtual and real. But result reveals that its performance is limited by the result of segmentation.If the segmentation techniques develop, our model would have better capacity. Future work can combine our proposed model with supervised models on real-world data to achieve better results.

\newpage

{\small
\bibliographystyle{ieee}
\bibliography{egbib}

\begin{thebibliography}{10}\itemsep=-1pt

\bibitem{abbeel2007application}
P.~Abbeel, A.~Coates, M.~Quigley, , and A.~Y. Ng.
\newblock An application of reinforcement learning to aerobatic helicopter.
\newblock flight. \emph{Advances in neural information processing systems},
  19:\penalty0 1, 2007.

\bibitem{CNNs}
A.~Alpher.
\newblock Frobnication.
\newblock {\em Journal of Foo}, 12(1):234--778, 2002.

\bibitem{EndtoEndLearningforSelf-DrivingCar}
A.~Alpher and J.~P.~N. Fotheringham-Smythe.
\newblock Frobnication revisited.
\newblock {\em Journal of Foo}, 13(1):234--778, 2003.

\bibitem{badrinarayanan2015segnet}
V.~Badrinarayanan, A.~Kendall, , and R.~Cipolla.
\newblock Segnet: A deep convolutional encoder-decoder architecture for image.
\newblock segmentation. \emph{arXiv preprint arXiv:1511.00561}, 2015.

\bibitem{nvidia}
M.~Bojarski, D.~D. Testa, D.~Dworakowski, B.~Firner, B.~Flepp, P.~Goyal, L.~D.
  Jackel, M.~Monfort, U.~Muller, J.~Zhang, X.~Zhang, J.~Zhao, , and K.~Zieba.
\newblock End to end learning for self-driving cars.
\newblock \emph{CoRR}, abs/1604.07316, 2016. URL
  \textup{http://arxiv.org/abs/1604.07316}.

\bibitem{dataset}
S.~Chen.
\newblock Autopilot-tensorflow, 2016.
\newblock URL \textup{https://github.com/SullyChen/Autopilot-TensorFlow}.

\bibitem{DBLP:journals/corr/CordtsORREBFRS16}
M.~Cordts, M.~Omran, S.~Ramos, T.~Rehfeld, M.~Enzweiler, R.~Benenson,
  U.~Franke, S.~Roth, , and B.~Schiele.
\newblock The cityscapes dataset for semantic urban scene understanding.
\newblock \emph{CoRR}, abs/1604.01685, 2016. URL
  \textup{http://arxiv.org/abs/1604.01685}.

\bibitem{endo2008learning}
G.~Endo, J.~Morimoto, T.~Matsubara, J.~Nakanishi, , and G.~Cheng.
\newblock Learning cpg-based biped locomotion with a policy gradient method:.
\newblock Application to a humanoid robot. \emph{The International Journal of
  Robotics Research}, 27\penalty0 (2):\penalty0 213--228, 2008.

\bibitem{gupta2017learning}
A.~Gupta, C.~Devin, Y.~Liu, P.~Abbeel, , and S.~Levine.
\newblock Learning invariant feature spaces to transfer skills with.
\newblock reinforcement learning. \emph{arXiv preprint arXiv:1703.02949}, 2017.

\bibitem{kohl2004policy}
N.~Kohl and P.~Stone.
\newblock Policy gradient reinforcement learning for fast quadrupedal.
\newblock locomotion. In \emph{Robotics and Automation, 2004. Proceedings.
  ICRA'04. 2004 IEEE International Conference on}, volume~3, pages 2619--2624.
  IEEE, 2004.

\bibitem{koutnik2013evolving}
J.~Koutn{\'\i}k, G.~Cuccu, J.~Schmidhuber, , and F.~Gomez.
\newblock Evolving large-scale neural networks for vision-based reinforcement.
\newblock learning. In \emph{Proceedings of the 15th annual conference on
  Genetic and evolutionary computation}, pages 1061--1068. ACM, 2013.

\bibitem{lillicrap2015continuous}
T.~P. Lillicrap, J.~J. Hunt, A.~Pritzel, N.~Heess, T.~Erez, Y.~Tassa,
  D.~Silver, , and D.~Wierstra.
\newblock Continuous control with deep reinforcement learning.
\newblock \emph{arXiv preprint arXiv:1509.02971}, 2015.

\bibitem{Long_2015_CVPR}
J.~Long, E.~Shelhamer, , and T.~Darrell.
\newblock Fully convolutional networks for semantic segmentation.
\newblock In \emph{The IEEE Conference on Computer Vision and Pattern
  Recognition (CVPR)}, June 2015.

\bibitem{A3C}
V.~Mnih, A.~P. Badia, M.~Mirza, A.~Graves, T.~P. Lillicrap, T.~Harley,
  D.~Silver, , and K.~Kavukcuoglu.
\newblock Asynchronous methods for deep reinforcement learning.
\newblock \emph{CoRR}, abs/1602.01783, 2016. URL
  \textup{http://arxiv.org/abs/1602.01783}.

\bibitem{mnih2015human}
V.~Mnih, K.~Kavukcuoglu, D.~Silver, A.~A. Rusu, J.~Veness, M.~G. Bellemare,
  A.~Graves, M.~Riedmiller, A.~K. Fidjeland, G.~Ostrovski, and et~al.
\newblock Human-level control through deep reinforcement learning.
\newblock \emph{Nature}, 518\penalty0 (7540):\penalty0 529--533, 2015.

\bibitem{pomerleau1989alvinn}
D.~A. Pomerleau.
\newblock Alvinn, an autonomous land vehicle in a neural network.
\newblock Technical report, Carnegie Mellon University, Computer Science
  Department, 1989.

\bibitem{DBLP:journals/corr/RusuVRHPH16}
A.~A. Rusu, M.~Vecerik, T.~Roth{\"{o}}rl, N.~Heess, R.~Pascanu, , and
  R.~Hadsell.
\newblock Sim-to-real robot learning from pixels with progressive nets.
\newblock \emph{CoRR}, abs/1610.04286, 2016. URL
  \textup{http://arxiv.org/abs/1610.04286}.

\bibitem{schulman2015trust}
J.~Schulman, S.~Levine, P.~Abbeel, M.~I. Jordan, , and P.~Moritz.
\newblock Trust region policy optimization.
\newblock In \emph{ICML}, pages 1889--1897, 2015.

\bibitem{DBLP:journals/corr/Shalev-ShwartzS16a}
S.~Shalev{-}Shwartz, S.~Shammah, , and A.~Shashua.
\newblock Safe, multi-agent, reinforcement learning for autonomous driving.
\newblock \emph{CoRR}, abs/1610.03295, 2016. URL
  \textup{http://arxiv.org/abs/1610.03295}.

\bibitem{tobin2017domain}
J.~Tobin, R.~Fong, A.~Ray, J.~Schneider, W.~Zaremba, , and P.~Abbeel.
\newblock Domain randomization for transferring deep neural networks from.
\newblock simulation to the real world. \emph{arXiv preprint arXiv:1703.06907},
  2017.

\bibitem{tzeng2016adapting}
E.~Tzeng, C.~Devin, J.~Hoffman, C.~Finn, P.~Abbeel, S.~Levine, K.~Saenko, , and
  T.~Darrell.
\newblock Adapting deep visuomotor representations with weak pairwise.
\newblock constraints. In \emph{Workshop on the Algorithmic Foundations of
  Robotics (WAFR)}, 2016.

\bibitem{wymann2000torcs}
B.~Wymann, E.~Espi{\'e}, C.~Guionneau, C.~Dimitrakakis, R.~Coulom, , and
  A.~Sumner.
\newblock Torcs, the open racing car simulator.
\newblock \emph{Software available at http://torcs. sourceforge. net}, 2000.

\bibitem{DBLP:journals/corr/YouPWL17}
Y.~You, X.~Pan, Z.~Wang, and C.~Lu.
\newblock Virtual to real reinforcement learning for autonomous driving.
\newblock {\em CoRR}, abs/1704.03952, 2017.

\bibitem{DBLP:journals/corr/ZhaoSQWJ16}
H.~Zhao, J.~Shi, X.~Qi, X.~Wang, and J.~Jia.
\newblock Pyramid scene parsing network.
\newblock {\em CoRR}, abs/1612.01105, 2016.

\end{thebibliography}
 }

	\end{document}